\documentclass[letterpaper]{article}
\usepackage{aaai2027}
\usepackage[hyphens]{url}
\usepackage{graphicx}
\usepackage{natbib}
\usepackage{caption}
\usepackage{algorithm}
\usepackage{algorithmic}

\usepackage{newfloat}
\usepackage{listings}
\DeclareCaptionStyle{ruled}{labelfont=normalfont,labelsep=colon,strut=off}
\floatstyle{ruled}
\newfloat{listing}{tb}{lst}{}
\floatname{listing}{Listing}

\usepackage{booktabs}

\newcommand{\ourmethod}{STARK}

\usepackage[utf8]{inputenc}
\usepackage[T1]{fontenc}
\usepackage{url}
\usepackage{tabularx}
\usepackage{array}
\usepackage{colortbl}
\usepackage{amsfonts}
\usepackage{nicefrac}
\usepackage{microtype}
\usepackage{xcolor}
\usepackage{amsmath}
\usepackage{amssymb}
\usepackage{multirow}
\usepackage{subcaption}
\usepackage[inline]{enumitem}
\colorlet{verylightgray}{lightgray!99!white}
\usepackage{framed}
\newenvironment{promptbox}[1]{\begin{framed}\noindent\textbf{#1}\par\smallskip\small\ttfamily}{\end{framed}}

\makeatletter\@ifundefined{iflatexml}{\newif\iflatexml\latexmlfalse}{}\makeatother
\title{STARK: Moving Towards Personalized Agents That Truly Learn and Remember}
\iflatexml
\makeatletter
\author{{Aryan Jain, Abhishek Jha*, Anish Nediyanchath, Promod Yenigalla}}
\lx@add@affiliations{RBS Tech Sciences, Amazon}
\lx@add@contact[role=affiliation]{\{arynjn, abhisjht, anishned, promy\}@amazon.com}
\lx@add@contact[role=note]{* Work done during internship}
\def\aaai@affiliations{}%
\makeatother
\else
\author {
    Aryan Jain,
    Abhishek Jha\protect\thanks{Work done during internship},
    Anish Nediyanchath,
    Promod Yenigalla
}
\affiliations {
    RBS Tech Sciences, Amazon\\
    \{arynjn, abhisjht, anishned, promy\}@amazon.com
}
\fi

\newcommand{\algwrap}[1]{\iflatexml\lxWithClass{alg_wrap}{#1}\else#1\fi}
\begin{document}

\maketitle

\begin{abstract}
Long-term memory is essential for personalized agentic systems that interact with users over long time periods, yet current architectures only capture what happened while leaving the reasoning of why it happened to be derived later by underlying LLMs.
 These events can, however, convey not only what a user experienced but also how that experience shaped their behavioral patterns.
We introduce \ourmethod{}, a dual-graph memory architecture which treats these events not just as nodes but also as edges between behavioral insights. \ourmethod{} maintains an Episodic Graph which records interactions as temporally, causally, and semantically linked episodes; and an Insight Graph which represents LLM-extracted behavioral signals scored by how many events support them, connected by the very events that led to them. At inference, behavioral knowledge is retrieved directly from insight nodes grounded by event-as-edges, eliminating repeated episodic traversal. The event-as-edge structure also naturally enables projecting the knowledge graph into a relational embedding space where cross-user transfer becomes possible under shared contextual circumstances, allowing the system to infer missing insights for cold-start users.
Evaluations show that \ourmethod{} improves by 1.48 judge score on LongMemEval information extraction and 0.93 on LoCoMo open domain questions over state of the art. On the public LongLaMP abstract-generation benchmark, cross-user transfer improves open-ended judge scores by 0.7 points and ROUGE-L by 3 at 5\% graph sparsity.
\end{abstract}

\section{Introduction}
\label{sec:introduction}

\begin{figure*}[t]
  \centering
  \includegraphics[width=\linewidth]{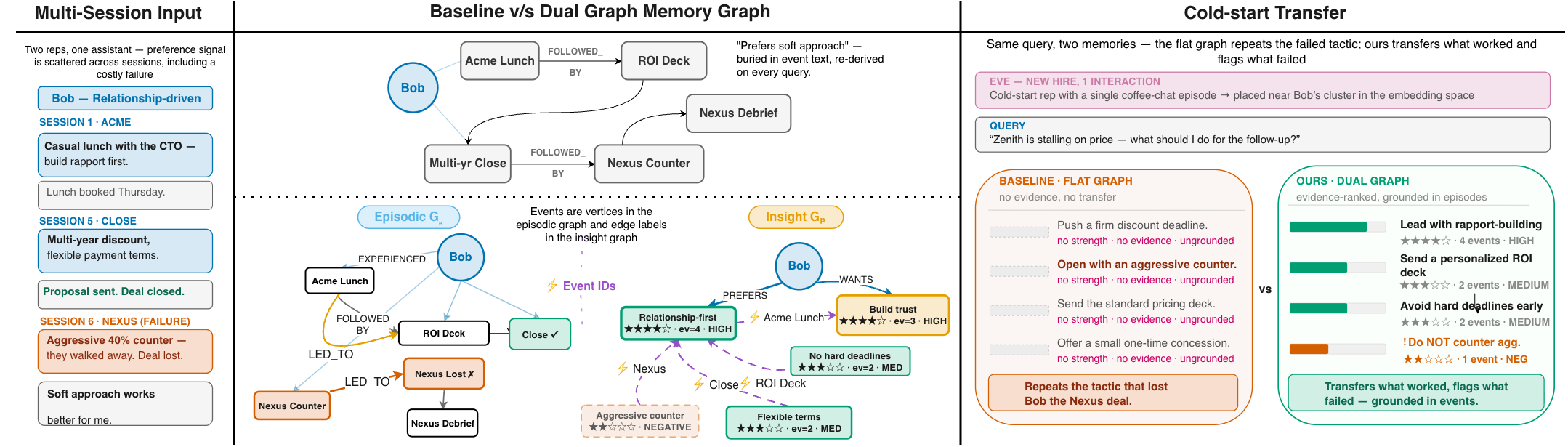}
  \caption{\textbf{From scattered multi-session dialogue to grounded cold-start personalization.} In the example shown, Bob tries different strategies with vendors ACME and NEXUS. The graph built for Bob, helps Eve (a new account manager) to close deal with their vendor using proven techniques.
  Baseline graph buries preferences while our \emph{dual-graph memory} separates episodes and insights (evidence-bearing nodes, strength shown as border thickness), with the \emph{same event as an edge label} bridging them.}
  \label{fig:stark_teaser_output}
\end{figure*}

As autonomous interactive agents are increasingly deployed to manage complex, long-running tasks such as workflow orchestration and support-ticket triage under evolving constraints ~\cite{autogen2023, surveyautoagent2023, casper}, their effectiveness relies on accumulating knowledge across long-horizon interactions. For example, for a deal negotiation agent, a compliance constraint surfaced during negotiation in one quarter must inform procurement strategy in the next; a failed negotiation strategy must not be repeated in the future. Moreover, many of these agents are domain-specific, wherein users pursue structurally similar goals and exhibit convergent behavioral patterns, making cross-user knowledge transfer a natural consequence of shared goal distributions.

Recent persistent-memory systems partially address this by constructing knowledge graphs from interactions \cite{zep, amem, jiang2025magma}, but maintain a single flat graph per user where events, preferences, constraints, and goals coexist as undifferentiated nodes. This design has two consequences. First, behavioral signals (the durable, reusable insights into how a user tends to act and decide, namely their preferences, constraints, goals, and values, distilled from many individual episodes and distinct from the episodes themselves) remain distributed across episodic nodes rather than being represented explicitly, which forces the LLM to re-derive them at query time, redundantly and often inconsistently \cite{richardson2023, hipporag}. Second, each user's graph is self-contained and provides no pathway for cross-user transfer. The common root cause is that these graphs capture \emph{what happened} while leaving the reasoning behind \emph{why it happened} to downstream inferencing.

We introduce \textbf{\ourmethod{}}, a dual-graph memory architecture built around this distinction. Our core contributions are as follows:

\begin{enumerate}
\item A \textbf{dual-graph architecture} that separates episodic events from behavioral knowledge. It is composed of an \textbf{Episodic Graph} with temporally, causally and semantically linked episodes and an \textbf{Insight Graph} that contains behavioral signals, each carrying evidence vectors from the episodes that support it. The two are joined by an \emph{event-as-edge bridge}: episodic identifiers serve as typed edge labels in the \textbf{Insight Graph}, keeping every insight grounded in the episodes that produced it.

\item A \textbf{cross-user transfer mechanism} for cold-start users. Insight nodes are embedded in a relational space where users with similar \emph{preference fingerprints} cluster together, and link prediction over that space infers a new user's likely signals without accessing another user's raw interactions.

\item We deploy \ourmethod{} for a production vendor-negotiation assistant and evaluate it in
an \textbf{8-week A/B pilot} ($N{=}40$ managers) cutting negotiation rounds per deal by $18\%$ and cycle time by $15\%$ ($p<0.05$).

\end{enumerate}

\section{Related Work}
\label{sec:related_work}

  \textbf{Memory-Augmented Generation:} RAG \cite{lewis2020retrieval} treats knowledge as a static corpus, and write-back memories (MemoryBank \cite{zhong2024memorybank}, Generative Agents \cite{park2023generative}, MemGPT
  \cite{packer2023memgpt}) persist interactions as flat text with no structure over the preferences, outcomes, or causal processes shaping behavior.
  Graph-based systems (GraphRAG \cite{edge2024graphrag}, Zep \cite{zep}, MAGMA \cite{jiang2025magma}, EverMemOS \cite{xu2025evermemos}, A-MEM \cite{amem}, Nemori \cite{nan2025nemori}) add episodic structure but record only what happened, treating behavioral signals as incidental nodes; even MAGMA's multiple graphs partition events rather than promoting signals to evidence-scored nodes, so neither links an insight to the experiences that formed it.

  \textbf{Preference Modeling:} Mem0 \cite{chhikara2025mem0} and MemoryOS \cite{kang2025memoryos} store weighted user facts, but RealPref \cite{realpref} and PAL-Bench \cite{palbench} show explicit extraction fails when preferences must be inferred from behavior. \ourmethod{} closes this gap with evidence-scored insight nodes bound to their forming episodes by an event-as-edge bridge.

\section{\ourmethod{} Framework}
\label{sec:approach}

\begin{figure*}[t]
\centering
\includegraphics[width=\linewidth]{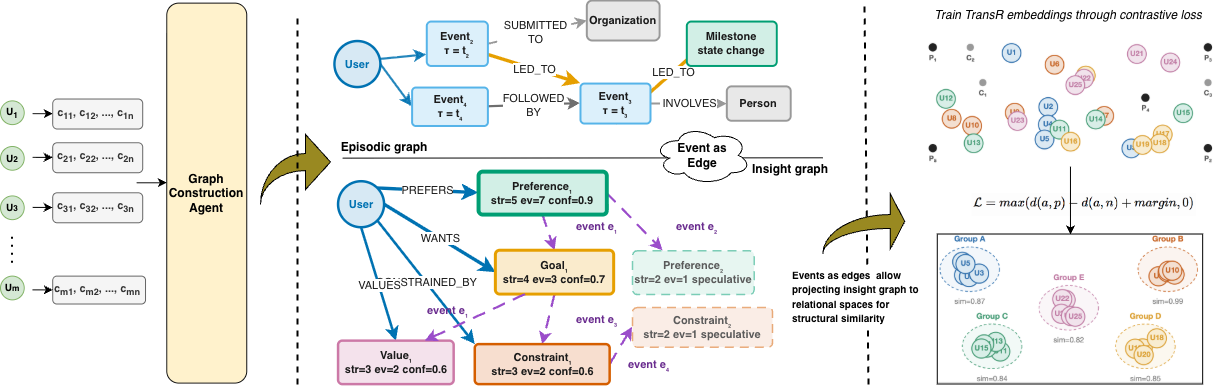}
\caption{Detailed architecture of STARK ingestion framework. All conversations are first ingested into the dual graphs bridged by episodic events which are then used for obtaining the relational embedding space enabling cross-user insight transfer.}
\label{fig:ingestion-arch}
\end{figure*}

Consider users $\mathcal{U} = \{u_1, \ldots, u_N\}$ interacting with an LLM agent over many sessions spanning long periods. For each $u_i$ the agent observes a temporally ordered stream of conversations $\mathcal{C}_i = \{c_1^i, \ldots, c_{T_i}^i\}$ of user-agent message pairs. The persistent-memory problem is to construct from $\mathcal{C}_i$ a structured memory $\mathcal{M}_i$ such that, given a query $q$ at any future step, the agent retrieves enough context to meaningfully advance the user's underlying goal, from acquiring information to a measurable outcome such as resolving a ticket or closing a deal. Figure~\ref{fig:ingestion-arch} illustrates the ingestion architecture, decomposing $\mathcal{M}_i = (\mathcal{G}_E^i, \mathcal{G}_I^i)$.

Fig.~\ref{fig:stark_teaser_output} shows how \ourmethod{} handles insights and facts about an account manager and assists them to negotiate deals.

\subsection{Episodic Graph}
\label{subsec:episodic_graph}

$\mathcal{G}_E^i = (V_E, E_E)$ captures the factual record of a user's interactions. Canonical node types are \{User, Event, Milestone, Organization\}, where Event nodes form the core, each a discrete interaction observed in conversation, and a Milestone marks a significant workflow or life-state change; the vocabulary is open, extended on demand by induced types such as Activity, Person, or Decision. Edges are temporal (\textsc{followed\_by}), causal (\textsc{led\_to}, \textsc{caused}), or action edges (domain-specific entity interactions, e.g.\ between users and the entities they act on, such as topics or artifacts), each carrying a timestamp $t(e)$. Together they support temporal, causal, and entity-centric graph retrieval. We also provide qualitative extraction examples on the LongLaMP abstract-generation domain in Appendix~\ref{appendix:qualitative}.

  \subsection{Insight Graph}
  \label{subsec:insight_graph}
  While $\mathcal{G}_E^i$ records what happened, $\mathcal{G}_I^i = (V_I, E_I)$ captures the behavior behind why it happened, with the $User^{i}$ node as the one fixed type anchoring each user to their signals.

  Edges $E_I$ carry two kinds of labels. Structural relations form a stable, closed set $R_s = \{\textsc{prefers}, \textsc{wants}, \textsc{values}, \textsc{tried}, \textsc{avoids}, \textsc{learned}\}$; a Preference \textsc{learned} a
  Constraint, for instance, is a signal-to-signal edge expressible only because signals are nodes rather than edge attributes. Event-linked
  relations $R_e = \{\mathrm{id}(e) : e \in V_E\}$ reuse episodic event identifiers as labels: an edge $(v_a, e_{id}, v_b) \in E_I$ means the relationship between $v_a$ and $v_b$ was observed during event $e_{id}$. This \emph{event-as-edge
  bridge} enables bidirectional navigation between a signal and its originating events, anchoring each insight to its episodic provenance without merging the two graphs (Fig.\ \ref{fig:ingestion-arch}).

  \textbf{Evidence vectors:} An agent must decide whether to \emph{act} on an inferred signal, and the cost is asymmetric: a one-off, weakly supported guess is risky, whereas a signal reinforced across many successful episodes warrants
  confident action. Each non-user signal $v \in V_I \setminus V_I^{User}$ (with $V_I^{User}$ the set of User nodes) therefore carries an evidence vector that lets retrieval rank signals by their support,
  \begin{equation}
  \label{eq:evidence-vector}
  ev(v) = \big(s(v),\, n(v),\, w(v),\, c(v)\big),
  \end{equation}
  updated whenever a conversation reinforces or modifies the signal. The \emph{evidence count} $n(v) \in \mathbb{N}^+$ is the number of distinct supporting episodes, and the \emph{outcome weight} $w(v) = \tfrac{1}{|R(v)|}\sum_{j \in R(v)} o_j \in [0,1]$ records how often acting on the signal achieved the user's goal, where $R(v)$ is the set of \emph{resolved} supporting episodes and each $o_j$ is set at ingest from in-conversation markers of whether acting on the signal succeeded, such as whether a proposed approach was adopted or well received. Unresolved or partial episodes still count toward $n(v)$ but are excluded from $w(v)$ rather than scored as failures, so pending evidence does not deflate the outcome weight; when no episode is yet resolved, $w(v){=}0.5$.
  \emph{Strength} grows with evidence and a consistency bonus, $s(v) = \sigma(\alpha_s \log_2(1+n(v)) + \lambda_s \mathbb{1}_{w(v)>\tau_w})$, with $\sigma$ the sigmoid, $\alpha_s$ the saturation rate, $\lambda_s$ the bonus magnitude, and
  $\tau_w$ the outcome-weight threshold; the log scaling yields diminishing returns. \emph{Confidence} rewards volume but penalizes inconsistent outcomes, $c(v) = \sigma(\alpha_c\, n(v) - \lambda_c\, w(v)(1-w(v)))$, with $\alpha_c$ scaling volume and $\lambda_c$ the variance penalty $w(1{-}w)$.

  \textbf{Contradictory signals:} Conflicting signals persist as separate nodes with independent evidence vectors rather than being resolved prematurely: a ``heavy data augmentation'' \textsc{prefers} signal that improved results in one subarea (high $w$) can coexist with an \textsc{avoids} signal learned after it hurt results on a different task (low $w$), disambiguated by the topic/subarea context on their bridging events. Retrieval surfaces both and resolves the conflict from this stored support, weighing evidence, query context, and recency rather than re-deriving it from raw history; when these leave the signals tied, \ourmethod{} abstains and presents the trade-off to the user.
\section{Cross-User Structural Transfer}
\label{sec:cross-user-transfer}

Each user's dual-graph is built independently, so every $(\mathcal{G}_E^i, \mathcal{G}_I^i)$ is a self-contained island: a sparse-history user has almost no signals to retrieve, exactly when guidance is most needed. Yet \ourmethod{} targets domain-specific settings whose users share a goal distribution and converge on recurring patterns. For example, among authors writing in the same research area, one author establishes that a particular method suits a class of problems, and the next author working on that problem class rediscovers the
  same relevant concepts from scratch. This is the same shared-goal structure that arises among account managers negotiating with a common vendor base in our deployment (Section~\ref{appendix:production}).

We therefore project every user's signal relations into one relational embedding space, learned from triples $(h, r, t)$ over the \emph{combined dual-graph}, the union of all users' $(\mathcal{G}_E^i, \mathcal{G}_I^i)$, and read transfer off the geometry by user-level similarity matching and link prediction over missing signals.

\textbf{Why a relation-aware embedding?}
Collaborative filtering \cite{koren2009matrix} collapses every interaction into one interacts-with edge, whereas our signals are heterogeneous: the same constraint node plays different geometric roles under \textsc{prefers} and under the collapsed \textsc{surfaced} relation. We therefore use TransR \cite{lin2015learning}, with entity embeddings $e \in \mathbb{R}^d$, relation embeddings $r \in \mathbb{R}^k$, and per-relation projections $M_r \in \mathbb{R}^{k \times d}$ that map entities into a relation-specific subspace before scoring $f(h, r, t) = \lVert M_r e_h + r - M_r e_t \rVert_2$. This projection, which a single-space model like TransE \cite{bordes2013translating} lacks, separates \textsc{prefers} from \textsc{surfaced} and drives clustering, drawing two users who connect to the same signal under the same relation together. It holds only among users sharing a domain and goal distribution, so we claim no cross-domain transfer. We further compare TransR against RotatE and CompGCN in Appendix~\ref{appendix:transr-comparison}. We use standard TransR notation: in a triple $(h,r,t)$, $t$ is the tail entity and $e$ an entity embedding, distinct from the episodic timestamp $t(e)$ and event $e$.

\textbf{Selective prefixing:}
The geometry produces transfer only if users share target nodes, so during triple extraction we prefix user-specific entities but leave behavioral-signal entities unprefixed, letting two users point at one shared signal node, the seam along which one user's geometry becomes informative for another (evidence in Appendix~\ref{app:design-justification}). Event-linked labels in $R_e$ appear mostly once, giving TransR no repeated examples to estimate $M_r$, so we collapse them to a single \textsc{surfaced} relation while keeping $R_s$ intact.

  \textbf{Preference fingerprints:}
  The first mechanism finds behaviorally similar users so a user, especially a cold-start one, can borrow their signals. Raw user-node embeddings fail, as a user node connects overwhelmingly to unshared episodic edges, so we represent each user only by their
  shared-signal edges, aggregated into a preference fingerprint:
  \begin{equation}
  \label{eq:fingerprint}
  \mathrm{fp}(u) = \frac{\sum_{(u,r,p)\in T_u} w_r\, e_p}{\bigl\lVert \sum_{(u,r,p)\in T_u} w_r\, e_p \bigr\rVert} ,
  \end{equation}
  where $T_u$ holds the triples connecting $u$ to shared signal entities and $w_r$ weights relation informativeness (\textsc{prefers} above \textsc{tried}). User similarity is the cosine between fingerprints, from which a user inherits candidate signals from its top-$m$ most similar users.

  \textbf{Entity-level link prediction:}
  Fingerprint matching needs a few existing signals to anchor the comparison, which cold-start users lack, so the second mechanism asks which specific signal the user is missing, scoring each candidate against its best-fitting link to what
  the user holds, $\mathrm{predicted}(u) = \arg\min_{p \notin P_{\text{existing}}} \min_{r \in R_p} f(\text{seed}, r, p)$, where seed iterates over $P_{\text{existing}}$ and $R_p \subset R_s$ are the preference-relevant relations obtained during graph construction. When no signals exist, the user node itself is the seed under \textsc{prefers} and \textsc{wants}, so a brand-new user still inherits common domain signals.

The dual-graph structure enables a retrieval strategy that queries both graphs with distinct objectives: the Insight Graph for evidence-ranked behavioral knowledge, and the Episodic Graph for narrative context and temporal grounding. Algorithm~\ref{alg:retrieval} describes how the retrieval system works and how context is assembled at the end of it.

\begin{algorithm}[t]
\caption{Dual-Graph Retrieval}
\label{alg:retrieval}
\iflatexml
\noindent\textbf{Require:} Query $q$, user graphs $(\mathcal{G}_E^i, \mathcal{G}_I^i)$, context budget $B$, optional TransR model\\
\textbf{Ensure:} Retrieved context $\mathcal{X}$
\par
\fi
\begin{algorithmic}[1]
\iflatexml\else
\REQUIRE Query $q$, user graphs $(\mathcal{G}_E^i, \mathcal{G}_I^i)$, context budget $B$, optional TransR model
\ENSURE Retrieved context $\mathcal{X}$
\fi
\STATE \algwrap{Extract entity mentions and query intent $\tau \in \{\textsc{behavioral}, \textsc{episodic}, \textsc{mixed}\}$ from $q$ via LLM, producing seed set $\mathcal{S}$}
\IF{TransR model available}
  \STATE Augment $\mathcal{G}_I^i$ with TransR entity predictions
  \STATE Retrieve similar-user insights via preference fingerprints
\ENDIF
\STATE \textbf{Insight retrieval:} \algwrap{Look up $\mathcal{S}$ in $\mathcal{G}_I^i$; retrieve behavioral signal nodes ranked by $\text{rank}(v) = s(v) \cdot c(v) \cdot (0.5 + w(v))$, so failure-derived signals ($w{=}0$) keep half weight}
\STATE \textbf{Episodic retrieval:} \algwrap{Look up $\mathcal{S}$ in $\mathcal{G}_E^i$; retrieve matching event nodes ranked by temporal recency and causal relevance}
\STATE \textbf{Bridge traversal:} \algwrap{For each retrieved behavioral signal, follow event-as-edge labels to $\mathcal{G}_E^i$; retrieve originating events as provenance}
\STATE \textbf{Context assembly:} \algwrap{Allocate budget $B$ according to intent $\tau$: behavioral signals first when $\tau = \textsc{behavioral}$, episodic content first when $\tau = \textsc{episodic}$, interleaved when $\tau = \textsc{mixed}$. Truncate lowest-ranked results to fit $B$. Cross-user signals are labeled separately.}
\STATE \textbf{return} $\mathcal{X}$
\end{algorithmic}
\end{algorithm}

\section{Experimental Setup}

We evaluate on five public benchmarks. LoCoMo~\cite{locomo} and LongMemEval~\cite{longmemeval} target long-horizon memory (exercising the Episodic Graph via temporal, causal, and multi-hop retrieval); RealPref~\cite{realpref} and PAL-Bench~\cite{palbench} are preference-oriented (exercising the Insight Graph). Evidence scoring is active only where reliable outcome labels exist (RealPref, LoCoMo, LongMemEval); on PAL-Bench and LongLaMP, whose data record no reception signal, signals are ranked uniformly. All four are single-user, so we evaluate cross-user transfer on the public LongLaMP abstract-generation benchmark~\cite{longlamp2024,salemi2024lamp}, whose authors act as users with disjoint train/test authors for a genuine cold-start setting. We report on the full public test split of each benchmark and simulate cold start by retaining $\{5-100\}\%$ of each user's graph operations (protocol in Appendix~\ref{appendix:eval-protocol}).

We compare against memory systems Mem0~\cite{chhikara2025mem0}, A-MEM~\cite{amem}, Zep~\cite{zep}, and MAGMA~\cite{jiang2025magma}, and full-context LLM baselines (Claude Haiku 4.5, Sonnet 4.5/4.6, and Opus 4.5). For fairness, every memory system and all three A/B arms use Claude Sonnet 4.6 as the backbone and the same $B{=}4000$-token retrieval budget as \ourmethod{} (Table~\ref{tab:hyperparameters}; full-context truncation policy in Appendix~\ref{appendix:eval-protocol}). MCQ, true/false, and ranking are scored by exact match and are therefore judge-independent; open-ended answers are scored $1$--$5$ ($\geq 4$ correct) by a GPT-5.1 judge.

\section{Results}
\label{sec:experiments_and_results}

\begin{table*}[t]
\centering
\small
\setlength{\tabcolsep}{1mm}
\begin{tabular}{@{}cccccccccccc@{}}
\toprule
\textbf{Model} &
  \multicolumn{4}{c}{\textbf{LoCoMo}} &
  \multicolumn{4}{c}{\textbf{LongMemEval}} &
  \multicolumn{3}{c}{\textbf{LongLaMP (abst gen.)}} \\ \midrule
 &
  \textbf{\begin{tabular}[c]{@{}c@{}}Single\\ Hop\end{tabular}} &
  \textbf{\begin{tabular}[c]{@{}c@{}}Multi\\ Hop\end{tabular}} &
  \textbf{Temporal} &
  \textbf{\begin{tabular}[c]{@{}c@{}}Open \\ Domain\end{tabular}} &
  \textbf{\begin{tabular}[c]{@{}c@{}}Information \\ extraction\end{tabular}} &
  \textbf{Abstention} &
  \textbf{Temporal} &
  \textbf{\begin{tabular}[c]{@{}c@{}}Knowledge\\ update\end{tabular}} &
  \textbf{RG-1} &
  \textbf{RG-L} &
  \textbf{\begin{tabular}[c]{@{}c@{}}Judge\\ (/5)\end{tabular}} \\ \midrule
\multicolumn{12}{c}{\textit{LLM-Only (Full Context)}} \\ \midrule
Claude Haiku 4.5   & 3.69 & 3.10 & 2.12 & 2.77 & 1.60 & 1.75 & 1.10 & 1.65 & 24.0 & 18.5 & 2.6 \\
Claude Sonnet 4.5  & 3.73 & 3.22 & 2.38 & 2.85 & 1.65 & 1.80 & 1.15 & 1.84 & 25.2 & 19.4 & 2.8 \\
Claude Sonnet 4.6  & 3.82 & 3.31 & 2.58 & 2.89 & 1.70 & 1.90 & 1.20 & 1.85 & 25.8 & 19.9 & 2.9 \\
Claude Opus 4.5    & \underline{3.84} & \underline{3.34} & \underline{2.63} & \underline{2.93} & \underline{1.77} & \underline{1.90} & \underline{1.25} & \underline{1.88} & \underline{26.5} & \underline{20.6} & \underline{3.0} \\
\midrule
\multicolumn{12}{c}{\textit{Long Term Memory Structures}} \\ \midrule
Mem0              & 2.90 & 2.70 & 2.30 & 3.10 & 2.10 & 2.25 & 1.55 & 2.20 & 28.0 & 21.5 & 3.3 \\
MAGMA              & 3.88 & 2.64 & 3.25 & 2.58 & 2.90 & 2.75 & 2.10 & 2.70 & 29.2 & 22.6 & 3.5 \\
A-MEM              & 2.75 & 2.50 & 2.10 & 2.40 & 1.90 & 2.00 & 1.40 & 1.95 & 26.8 & 20.8 & 3.1 \\
Zep                & 2.80 & 2.55 & 2.20 & 3.15 & 2.00 & 2.15 & 1.50 & 2.10 & 28.5 & 22.0 & 3.4 \\
\textbf{\ourmethod{}} & \textbf{3.98} & \textbf{3.56} & \textbf{3.77} & \textbf{4.08} & \textbf{4.38} & \textbf{2.81} & \textbf{2.62} & \textbf{3.32} & \textbf{31.5} & \textbf{24.8} & \textbf{4.7} \\
\bottomrule
\end{tabular}%
\caption{Comparison of \ourmethod{} against baselines on LoCoMo, LongMemEval, and LongLaMP. All LoCoMo and LongMemEval judge scores are on a $1$--$5$ scale. \ourmethod{} achieves best LoCoMo single-hop (3.98), multi-hop (3.56), temporal reasoning (3.77), and open-domain (4.08), as well as best LongMemEval info extraction (4.38) and temporal reasoning (2.62).}
\label{tab:master-table-1A}
\end{table*}
\begin{table}[t]
\centering
\small
\setlength{\tabcolsep}{1.5mm}
\begin{tabular}{@{}lccccccccc@{}}
\toprule
\multirow{4}{*}{\textbf{Model}} & \multicolumn{5}{c}{\textbf{RealPref}} & \multicolumn{3}{c}{\textbf{PAL-Bench}} \\ \cmidrule(lr){2-6} \cmidrule(lr){7-9}
& \textbf{O.E.} & \multicolumn{3}{c}{\textbf{MCQ (\%)}} & \textbf{T/F} & \textbf{R.S.} & \multicolumn{2}{c}{\textbf{Rank (\%)}} \\ \cmidrule(lr){3-5} \cmidrule(lr){8-9}
& \textbf{(/5)} & \textbf{Ex.} & \textbf{Ge.} & \textbf{To.} & \textbf{(\%)} & \textbf{(/5)} & \textbf{T2} & \textbf{B2} \\
\midrule
\multicolumn{9}{c}{\textit{LLM-Only (Full Context)}} \\ \midrule
Haiku 4.5   & 2.03 & 75.2 & 70.7 & 74.2 & 54.9 & 2.69 & 28.6 & 34.8 \\
Sonnet 4.5  & 2.07 & 84.3 & 82.5 & 83.9 & 57.2 & 2.73 & 33.8 & 37.0 \\
Sonnet 4.6  & 2.09 & 85.0 & 82.0 & 84.3 & 61.5 & 2.77 & 32.6 & 36.3 \\
Opus 4.5    & 2.15 & 85.4 & 84.0 & 85.1 & 64.1 & 2.81 & 34.0 & 37.7 \\ \midrule
\multicolumn{9}{c}{\textit{Long Term Memory Systems}} \\ \midrule
Mem0       & 4.02 & 92.6 & 89.6 & 91.9 & 78.9 & 3.87 & 37.7 & 38.8 \\
MAGMA       & 3.80 & 94.0 & 91.5 & 93.4 & 80.0 & 3.70 & 40.0 & 42.0 \\
A-MEM       & 3.50 & 89.0 & 85.0 & 88.1 & 74.0 & 3.40 & 35.0 & 36.0 \\
Zep         & 3.90 & 91.0 & 88.0 & 90.3 & 77.0 & 3.60 & 36.0 & 37.5 \\
\textbf{\ourmethod{}} & \textbf{4.19} & \textbf{98.4} & \textbf{97.3} & \textbf{98.1} & \textbf{86.5} & \textbf{4.03} & \textbf{41.3} & \textbf{44.2} \\
\bottomrule
\end{tabular}
\caption{Comparison of \ourmethod{} against baselines on RealPref and PAL-Bench. Ex./Ge./To. = MCQ accuracy on RealPref's original/generalized/all, macro-averaged over users; T2/B2 = share of the top/bottom 2 ranked candidates; O.E. = open-ended judge; R.S. = requirement-restatement score.}
\label{tab:master-table-1B}
\end{table}
\begin{table}[t]
\centering
\small
\setlength{\tabcolsep}{1mm}
\begin{tabular}{@{}lcccccc@{}}
\toprule
\textbf{Method} & \textbf{.05} & \textbf{.1} & \textbf{.2} & \textbf{.3} & \textbf{.5} & \textbf{1.0} \\
\midrule
No-matching       & 1.3/12 & 2.0/14 & 2.7/16 & 3.6/18 & 4.0/19 & 4.4/20 \\
Random-matching    & 1.7/12 & 2.3/14 & 3.1/16 & 3.8/18 & 4.2/20 & 4.4/21 \\
Label-matching     & 1.8/13 & 2.4/15 & 3.3/17 & 4.0/19 & 4.3/21 & 4.5/22 \\
TransR(user)      & 1.8/13 & 2.5/15 & 3.4/18 & 4.1/20 & 4.4/22 & 4.6/23 \\
TransR(entity)    & 2.0/14 & 2.6/16 & 3.5/18 & 4.3/21 & 4.5/22 & 4.6/24 \\
TransR(combined)  & 2.0/15 & 2.7/16 & 3.7/19 & 4.4/21 & 4.6/23 & 4.7/25 \\
\bottomrule
\end{tabular}
\caption{Cross-user transfer impact on Open-Ended score (/5) / ROUGE-L at varying graph sparsity levels on the public LongLaMP abstract-generation benchmark. TransR combined = fingerprint similarity + entity link prediction.}
\label{tab:cross-user-transf-ablations}
\end{table}

\begin{figure}[t]
\centering
\includegraphics[width=\linewidth]{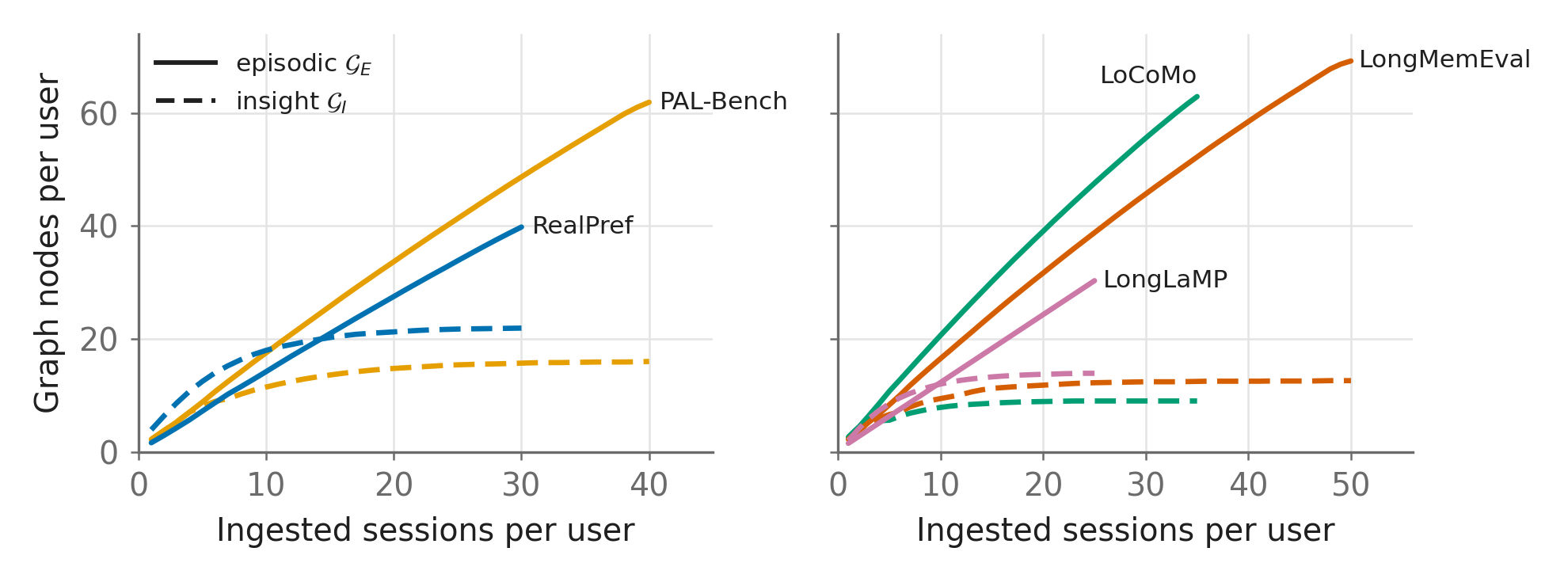}
\caption{Per-dataset graph growth with ingested sessions. Episodic nodes $\mathcal{G}_E$ (solid) grow roughly linearly; insight nodes $\mathcal{G}_I$ (dashed) saturate as recurring signals merge in place, keeping the retrieved surface small as history accumulates.}
\label{fig:stark-scaling-plot}
\end{figure}

\textbf{Graph structure matches full-context on short histories and dominates on long ones.} On LoCoMo (Table~\ref{tab:master-table-1A}, scores on a $1$--$5$ scale), \ourmethod{} is best across all four categories: it leads on single-hop (3.98 vs.\ MAGMA: 3.88, $+0.10$; vs.\ best LLM-only Opus 4.5: 3.84) and substantially outperforms all baselines
on multi-hop (3.56 vs.\ MAGMA: 2.64, $+0.92$), temporal reasoning (3.77 vs.\ MAGMA: 3.25, $+0.52$), and open-domain queries (4.08 vs.\ Zep: 3.15, $+0.93$). Notably, on LoCoMo
every other memory system does worse than full-context Opus~4.5 on multi-hop ($2.50$--$2.70$ vs.\ $3.34$): lossy retrieval hurts when the history already fits, and \ourmethod{} is the only system that avoids this cost. LongMemEval is the opposite case: histories no longer fit and full context collapses (information extraction around $1.6$--$1.8$),
so structured retrieval becomes essential. Here \ourmethod{} achieves 4.38 on information extraction ($+1.48$ over MAGMA), 2.62 on temporal reasoning ($+0.52$), and 3.32 on knowledge update ($+0.62$). On LongMemEval abstention we match MAGMA, as our retrieval passes nodes similar to the query entities and biases the model toward answering.

\textbf{Explicit preference nodes drive performance gains on preference-oriented datasets.} On RealPref (Table~\ref{tab:master-table-1B}), the gain concentrates where memory is irreplaceable: the open-ended score, where the model must produce the preference itself, jumps $+2.04$ over LLM-only ($2.15\!\to\!4.19$), whereas MCQ, where the answer is one of the given options, rises far less ($+13.0$pp, to $98.1\%$). Both gains share one source: preferences exist as evidence-scored nodes retrieved and ranked directly by support and consistency, not re-derived from raw text at query time. This ranking is what free-text stores cannot match: Mem0 nearly matches our open-ended fluency ($4.02$ vs.\ $4.19$) yet trails on MCQ and T/F ($91.9\%$, $78.9\%$ vs.\ $98.1\%$, $86.5\%$), and \ourmethod{} has the smallest Explicit-to-Generalized MCQ drop of any system ($1.1$pp vs.\ $1.4$--$4.5$pp), showing the explicit structure also helps infer unstated preferences. PAL-Bench isolates this ranking step by removing it: it has no action-outcome signal, so \ourmethod{} skips evidence scoring and treats signals uniformly. Its small residual edge (Top-2 $41.3\%$ vs.\ MAGMA $40.0\%$) therefore comes
from the dual-graph structure alone, and the low scores across all methods reflect how hard it is to order preferences with no success signal to rank on.

\begin{table*}[t]
\centering
\small
\setlength{\tabcolsep}{1mm}
\begin{tabular}{@{}lcccccc@{}}
\toprule
\textbf{Dataset} &
\textbf{Full System} &
\textbf{W/o TransR} &
\textbf{W/o $G_I$} &
\textbf{W/o Evidence} &
\textbf{W/o $G_E$} &
\textbf{W/o Graph} \\ \midrule
LoCoMo (Single-Hop)      & 3.98 / 1786 & 3.98 / 1786 & 3.90 / 1400 & 3.95 / 1786 & 3.55 / 1350 & 3.82 / 2466 \\
LongMemEval (Info extraction) & 4.38 / 1839 & 4.38 / 1839 & 4.05 / 1500 & 4.30 / 1839 & 2.60 / 1400 & 1.70 / 3823 \\
LongLaMP (Open ended)    & 4.70 / 1900 & 4.41 / 1810 & 4.05 / 1500 & 4.70 / 1900 & 4.20 / 1400 & 2.90 / 4500 \\
RealPref  (Open ended)   & 4.19 / 5531 & 4.19 / 5531 & 2.95 / 4800 & 3.70 / 5531 & 3.85 / 4500 & 2.09 / 3388 \\
PAL-Bench (R.S.)   & 4.03 / 7295 & 4.03 / 7295 & 3.10 / 6200 & 4.03 / 7295 & 3.75 / 5800 & 2.77 / 7242 \\
\bottomrule
\end{tabular}%
\caption{Full system performance when removing one component at a time. Each cell: \ourmethod{} score (1--5) / response generation latency (ms) with LLM backbone as Claude Sonnet 4.6. Evidence scoring is disabled on PAL-Bench and LongLaMP, which have no outcome signal; TransR affects only LongLaMP, the only shared-domain benchmark.}
\label{tab:stark-comp-ablations}
\end{table*}

\begin{figure}[t]
\centering
\includegraphics[width=\linewidth]{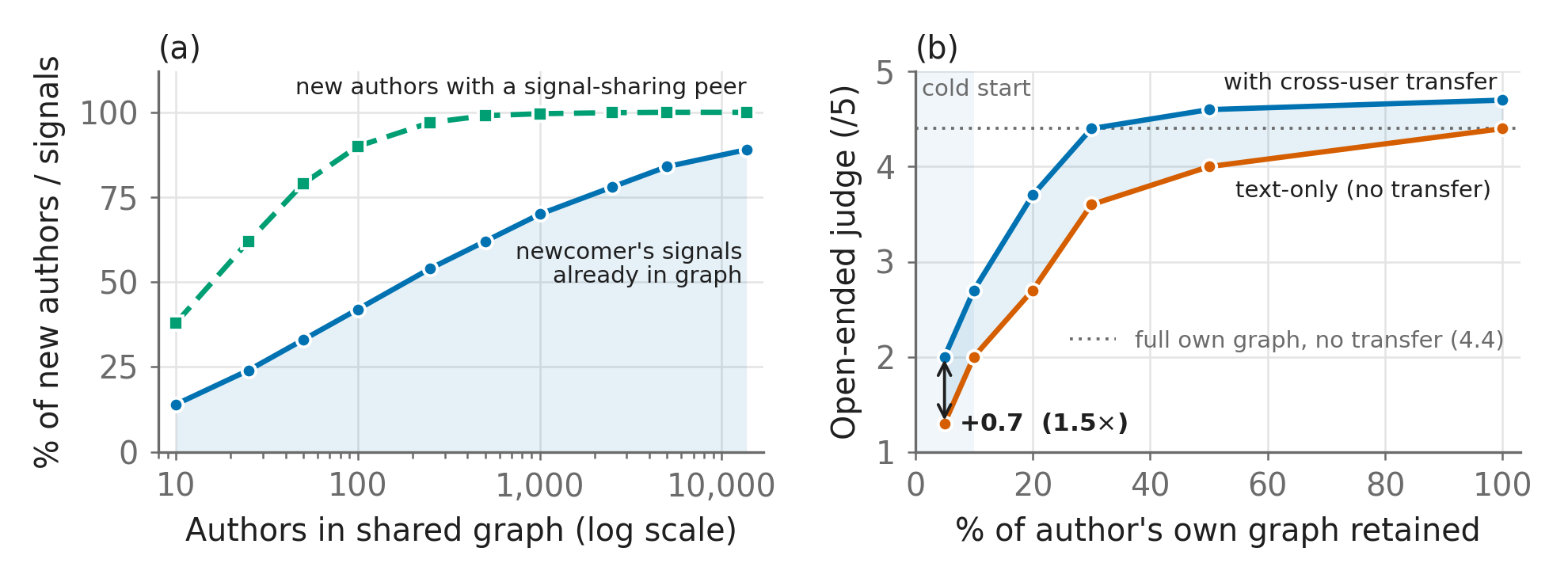}
\caption{Cross-user transfer scales on both deployment axes. (a) With more users, more of a newcomer's signals already exist in the collective graph (blue) and a signal-sharing peer becomes near-certain (green). (b) At 5\% of their own history, transfer recovers 2.0/5 vs.\ 1.3/5 text-only (1.5$\times$).}
\label{fig:transfer-scaling}
\end{figure}

\textbf{TransR enables 1.5$\times$ quality recovery at extreme cold-start.} At 5\% graph sparsity (Table~\ref{tab:cross-user-transf-ablations}, Figure~\ref{fig:transfer-scaling}b), TransR combined achieves 2.0/5 and 15 ROUGE-L vs.\ 1.3/12 for text-only. This is where transfer helps most: in a shared-goal domain most of a new user's signals already exist in the collective graph, and coverage grows with the user base (Fig.~\ref{fig:transfer-scaling}a).
Entity-level link prediction dominates across all sparsity levels; combining it with fingerprint similarity adds a further gain. We however see that random matching almost always performs better than no-matching. This means in a shared domain, cross-user insights are prevalent and useful.

\textbf{Each graph matters only for its own kind of query.} Removing the \textbf{Episodic} Graph badly hurts the factual and temporal benchmarks (LongMemEval $4.38\!\to\!2.60$; LoCoMo $-0.43$) but barely affects preferences (RealPref $-0.34$); removing the \textbf{Insight} Graph does the opposite (RealPref $4.19\!\to\!2.95$; PAL-Bench $-0.93$). Evidence scoring matters most for preferences (RealPref $-0.49$) and only marginally for factual recall (LoCoMo $-0.03$, LongMemEval $-0.08$), whose answers come mainly from the Episodic Graph; on LongLaMP and PAL-Bench, whose data carry no reception signal, it is disabled, so the two columns coincide. TransR affects only the shared-domain LongLaMP ($-0.29$, $+90$\,ms at $O(1)$). On long histories \ourmethod{} is also faster than full context (e.g.\ LongMemEval $1839$ vs.\ $3823$\,ms) as it sends a small subgraph, while insights saturate as episodes grow (Fig.~\ref{fig:stark-scaling-plot}), keeping the read path flat.

\section{Deployment and Impact}
\label{appendix:production}

\ourmethod{} is deployed as a drop-in \emph{memory layer} for a \textbf{vendor-negotiation assistant} used by Account Managers (AMs) who negotiate terms with thousands of vendors. Negotiations span many sessions across quarters, constraints surface mid-deal and must inform later ones, and closing strategies sit buried in each AM's history; new hires start cold, and the team rediscovers the same constraints independently.

We ran an $8$-week live A/B test over the $40$ AMs ($18$ new hires under $6$ months, $22$ with $3$+ years). It is paired at the vendor level: each of a manager's five vendors gets three otherwise identical deals, run separately to avoid carryover: memory-on, memory-on with cross-user transfer disabled (transfer-off), and memory-off control. This yields $200$ vendor triples ($600$ deals), with the order of the three arms randomized within each triple, and the transfer-off arm isolates cross-user transfer with the rest of the memory layer fixed. Each AM's dual-graph was initialized from up to a year of their prior history (under six months for new hires), so every arm retrieves over the AM's full available history from the first deal; the $8$-week window scopes only the deals conducted in the pilot, all $600$ of which closed, so cycle time carries no closure-based selection. Vendors respond live, each deal carries an AM-provided target, and all arms use Claude Sonnet 4.6. Memory-on cut negotiation rounds per deal by $18\%$ and cycle time by $15\%$ (paired $t$-test, $n{=}200$, both $p<0.05$), raised deals closed at or above target by $9$\,pp, and lifted blind guidance ratings by $0.8$/$5$. The gains concentrate in new hires ($\sim$$24\%$/$\sim$$19\%$ vs.\ experienced $\sim$$13\%$/$\sim$$11\%$). The transfer-off arm pins this on cross-user transfer: without it, memory-on retained only $\sim$$11\%$/$\sim$$9\%$, so transfer supplies roughly $40\%$ of the gain overall and most of it for new hires (new $\sim$$9\%$ vs.\ $\sim$$24\%$; experienced $\sim$$12\%$ vs.\ $\sim$$13\%$).
Through the pilot we learned that the AMs preferred inferred signals \emph{surfaced} as accept-or-dismiss options rather than acted on, so we leave the decision to them. We also observed that new--experienced fingerprint similarity (Eq.~\ref{eq:fingerprint}) rose $0.31\!\to\!0.58$, a descriptive within-arm trend (not isolated by the transfer-off arm).
\paragraph{Why it works?} Surfacing known constraints and concession patterns up front removes rounds spent rediscovering them; outcome-weighting (Eq.~\ref{eq:evidence-vector}) front-loads previously successful tactics and down-ranks rejected ones, shortening the cycle; and for new hires this knowledge arrives via cross-user transfer. Cost, governance, and rollout are in Appendix~\ref{app:deployment-ops}.

\section*{Conclusion}

We presented \ourmethod{}, a dual-graph memory that decouples episodic events from behavioral knowledge via an event-as-edge bridge, enabling evidence-ranked retrieval and cross-user transfer for cold-start inference.
However, \ourmethod{} has several important limitations. TransR updates in $O(1)$ but its per-relation projections have fixed capacity, so embedding quality degrades as the graph densifies. Privacy is enforced agentically, not formally: we do not bound residual leakage of aggregate fingerprints under membership inference. Transfer is domain-specific, holding only among users who share a goal distribution.
In future work, we plan to use the graph to look forward rather than back. Using behavioral paths as a search tree and running an MCTS-style rollout over likely trajectories can help the agent act on the path most likely to reach the user's goal.

\bibliography{aaai2027}

\appendix
\section{Appendix}

\subsection{Hyperparameter setup}
\label{appendix:hyperparam-setup}

Table~\ref{tab:hyperparameters} consolidates the hyperparameters of \ourmethod{} that are free rather than fixed by the architecture, grouped by the component that uses them. The evidence-scoring constants govern how strength and confidence saturate with accumulated evidence; the retrieval constant sets the read-path context budget (Algorithm~\ref{alg:retrieval}); and the transfer constants configure neighbor selection, fingerprint weighting, and the TransR embedding.
All remaining settings follow standard practice.

The evidence-scoring constants are set so that strength rises smoothly from $\approx\!0.67$ at a single supporting episode to $\approx\!0.94$ at fifteen, with the consistency bonus applied once a signal's outcome weight clears the majority threshold $\tau_w = 0.5$. The fingerprint weights $w_r$ rank structural relations by how strongly they express a durable preference: $w_{\textsc{prefers}} = w_{\textsc{wants}} = 1.0$, $w_{\textsc{values}} = 0.8$, $w_{\textsc{learned}} = 0.7$, $w_{\textsc{avoids}} = 0.6$, and $w_{\textsc{tried}} = 0.5$.

\begin{table}[ht]
\centering
\begin{tabular}{@{}llc@{}}
\toprule
Symbol & Description & Value \\
\midrule
\multicolumn{3}{@{}l}{\textit{Evidence scoring }} \\
$\alpha_s$ & strength saturation rate & $0.7$ \\
$\lambda_s$ & consistency-bonus magnitude & $0.5$ \\
$\tau_w$ & outcome-weight threshold & $0.5$ \\
$\alpha_c$ & confidence volume sensitivity & $0.5$ \\
$\lambda_c$ & outcome-variance penalty & $2.0$ \\
\midrule
\multicolumn{3}{@{}l}{\textit{Retrieval (Algorithm~\ref{alg:retrieval})}} \\
$B$ & context budget (tokens) & $4000$ \\
\midrule
\multicolumn{3}{@{}l}{\textit{Cross-user transfer}} \\
$m$ & similar users per user & $5$ \\
\midrule
\multicolumn{3}{@{}l}{\textit{TransR embedding }} \\
$d$ & entity embedding dimension & $64$ \\
$k$ & relation embedding dimension & $64$ \\
$\gamma$ & ranking-loss margin & $1.0$ \\
$\eta$ & learning rate & $0.01$ \\
-- & training epochs & $100$ \\
-- & negatives per positive triple & $1$ \\
\bottomrule
\end{tabular}
\caption{Hyperparameter settings for \ourmethod{}, grouped by the component that uses them. Evidence-scoring constants are set by design, not tuned; $m$ is tuned on the LongLaMP validation split, whose authors are disjoint from the $4{,}560$ test authors; TransR and retrieval-budget values follow standard practice.}
\label{tab:hyperparameters}
\end{table}

\subsection{LongLaMP Abstract-Generation Benchmark}
\label{appendix:internal-dataset}
\label{app:internal-dataset}

We evaluate on the public LongLaMP abstract-generation benchmark~\cite{longlamp2024,salemi2024lamp} (the \texttt{abstract\_generation\_user} split), whose users are authors drawn from a large citation-network corpus. The split places disjoint authors in train and test ($13{,}693$ train / $4{,}560$ test, zero overlap), so every test author is genuinely unseen---a true cold-start setting for cross-user transfer. A separate validation split, whose authors are disjoint from the test authors, is used only for hyperparameter tuning (Appendix~\ref{appendix:hyperparam-setup}). Each author's profile is their set of prior paper abstracts, and the personalization target is the abstract of a held-out paper.

We treat each author as a user, each prior paper as an episode in the Episodic Graph, and the research topics, methods, and concepts recurring across an author's abstracts as the behavioral-signal nodes of the Insight Graph. Because authors in the same research area converge on shared topics and methods, these signal nodes recur across users, providing the shared-domain structure that cross-user transfer relies on. For the clustering metric of Appendix~\ref{app:design-justification} we group authors by research field (assigned per author by an LLM from their profile abstracts) to obtain the ground-truth grouping.

\subsection{Evaluation Protocol}
\label{appendix:eval-protocol}

We report on the full public test split of each benchmark: LoCoMo's $10$ conversations, LongMemEval-S's $500$ questions, RealPref's $100$ users and $1{,}300$ preferences, and PAL-Bench's $100$ users; the LongLaMP split is detailed above (Appendix~\ref{appendix:internal-dataset}). Cold start is simulated by retaining $\{5,10,20,30,50,100\}\%$ of each user's graph operations to measure resilience to sparsity; at low sparsity we exclude factual MCQ and T/F questions whose answers are absent from the sparse graph. Evidence scoring uses per-episode outcome labels $o_j$ inferred at ingest by an LLM on RealPref, LoCoMo, and LongMemEval, where in-conversation reception signals exist; PAL-Bench and LongLaMP carry no such signal, so evidence scoring is disabled and their signals are ranked uniformly. Every memory system retrieves under the same $B{=}4000$-token budget as \ourmethod{}; the full-context LLM baselines receive the entire history when it fits the backbone's context window and, on histories that exceed it, are truncated to the most recent tokens that fit.

\subsection{TransR training details}
\label{appendix:transr-training}

We train a single TransR model over the combined multi-user graph, with entity embeddings $e \in \mathbb{R}^{d}$, relation embeddings $r \in \mathbb{R}^{k}$, and per-relation projection matrices $M_r \in \mathbb{R}^{k \times d}$, scored by $f(h,r,t) = \lVert M_r e_h + r - M_r e_t \rVert_2$. Training minimizes a margin-based ranking loss over the positive triples in the combined set against tail-corrupted negatives (a positive $(h, r, t)$ paired with $(h, r, t')$ for a sampled $t' \neq t$), pushing the geometry to score observed structure below corrupted structure; we use otherwise standard TransR hyperparameters and do not tune the embedding dimensions $d$ and $k$ per dataset.
A full offline train takes $58$\,s on the LongLaMP training set. Between full retrains the model is updated incrementally in $O(1)$ per new triple as conversations arrive, and a periodic full retrain admits new users and entities; in the interval before a user appears in a retrain, cold-start link prediction serves them off the existing embedding space. We adopt TransR because a live memory needs incremental, no-full-graph-recompute updates, which rules out CompGCN (it requires full message-passing at inference); we prefer it to RotatE, which is competitive on our metrics, for its interpretable per-relation subspaces (see Appendix~\ref{appendix:transr-comparison}).

\subsection{TransR comparison with other embedding models}
\label{appendix:transr-comparison}

The Insight Graph contains semantically heterogeneous relations (\textsc{prefers}, \textsc{surfaced}, \textsc{wants}) that require distinct geometric subspaces: the space in which a user \emph{prefers} a behavioral pattern differs from the space in which that pattern \emph{surfaces} a constraint. TransR addresses this directly via per-relation projection matrices $\mathbf{M}_r$ that map entities into relation-specific subspaces before translational scoring ($\mathbf{M}_r \mathbf{h} + \mathbf{r} \approx \mathbf{M}_r \mathbf{t}$). RotatE~\cite{sun2019rotate} models relations as rotations in a single complex space; empirically it clusters comparably to TransR (Table~\ref{tab:design-embedding}), so we claim no quality advantage over it. We nonetheless adopt TransR because its explicit per-relation projection matrices expose an inspectable subspace per relation, which suits the provenance and audit requirements of the deployment, and because its real entity space matches the fingerprint's real-vector aggregation $\mathrm{fp}(u) = \sum w_r \cdot \mathrm{emb}(p_i)$. CompGCN~\cite{vashishth2020compgcn} entangles entity representations across relation types during neighborhood aggregation and requires full message-passing at inference, making it incompatible with the $O(1)$ per-session incremental updates a live memory needs; TransR and RotatE both support such updates, CompGCN does not.

\subsection{Deployment: Cost, Governance, and Rollout}
\label{app:deployment-ops}

\paragraph{Engineering readiness.} \ourmethod{} concentrates cost on the one-time write path and keeps the read path flat as a user's graph grows. Ingestion is the cost center: each conversation is processed by two sequential LLM passes (episodic then insight), and a repeated behavioral signal is folded into an existing node by recomputing its evidence vector in place (Eq.~\ref{eq:evidence-vector}) rather than recomputing the graph, so write cost is bounded by the two passes and is independent of accumulated graph size. The read path is a fixed-budget filter over evidence-ranked signals: a query assembles about $3.8$K tokens from about $74$ entities in $710$--$770$\,ms regardless of how many sessions the user has logged, with cross-user transfer adding only about $90$\,ms. The embedding component is similarly write-heavy and read-cheap: TransR is trained offline ($58$\,s on the LongLaMP training set used in our experiments) and served with $O(1)$ incremental updates per new triple, with periodic full retraining ($\sim\!11$\,min at $10$K users) to admit new users and entities; between retrains, cold-start link prediction serves users the model has not yet seen. Table~\ref{tab:prod-cost} consolidates the cost and governance profile.

\paragraph{Governance, privacy, and compliance.} Cross-user transfer shares only the behavioral-signal layer: the LLM-extracted names and short descriptions of unprefixed signal nodes, plus the learned embeddings and fingerprint aggregates over them. Episodic records, raw conversation text, and user-specific entities are user-prefixed and never cross a user boundary, and transferred signals are rendered under explicit caveat headers kept separate from a user's own memory. Erasure maps directly onto the \texttt{DELETE\_NODE} and \texttt{UNLINK} operations, with a forgotten user's contribution to the shared embedding clearing at the next full retrain. The event-as-edge bridge keeps every acted-on signal traceable to its supporting episodes, giving the provenance an enterprise audit requires. These properties cleared the security and privacy sign-off required before the pilot went live.

\begin{table}[t]
\centering
\small
\resizebox{\columnwidth}{!}{%
\begin{tabular}{lll}
\toprule
Phase & Metric & Value \\
\midrule
\multirow{2}{*}{Ingestion}
  & Passes per conversation & $2$ (episodic, insight) \\
  & Latency / tokens per conv. & $6.2$\,s / $4.3$K \\
\midrule
\multirow{2}{*}{Retrieval}
  & Entities / tokens / latency & $\sim$$74$ / $\sim$$3.8$K / $710$--$770$\,ms \\
  & Cross-user add-on & $+90$\,ms \\
\midrule
\multirow{2}{*}{Embedding}
  & Full train (LongLaMP) & $58$\,s \\
  & Incremental / retrain at scale & $O(1)$ / $\sim$$11$\,min @ $10$K \\
\midrule
\multirow{2}{*}{Governance}
  & Cross-user channel & signal nodes + embeddings \\
  & Forget / PII policy & \texttt{DELETE\_NODE} / redaction \\
\bottomrule
\end{tabular}
}
\caption{Production cost and governance profile for \ourmethod{}. Cost concentrates on the one-time write path while the read path stays cheap and graph-size-stable.}
\label{tab:prod-cost}
\end{table}

\paragraph{Rollout plan.} We advanced through four staged gates (Table~\ref{tab:rollout}), each with an explicit go/no-go criterion, so that no stage advanced on academic proxies alone: offline replay on historical logs, shadow mode with recommendations hidden from managers, the A/B pilot, and general availability. Stages~0--2 have passed their gates; on the strength of the pilot we are now beginning the Stage~3 rollout to all managers, with scheduled retraining, monitoring, and rollback in place.

\begin{table}[t]
\centering
\small
\resizebox{\columnwidth}{!}{%
\begin{tabular}{llc}
\toprule
Stage & Go/no-go gate & Status \\
\midrule
0. Offline replay & Recover $\geq 80\%$ known constraints; & Passed \\
 & judge uplift $\geq +0.5$ & \\
\midrule
1. Shadow & p95 read $\leq 900$\,ms; $0$ PII leaks; & Passed \\
 & $\geq 90\%$ traceable & \\
\midrule
2. Pilot A/B & $\geq 15\%$ fewer rounds; & Passed \\
 & positive blind guidance-rating lift & \\
\midrule
3. GA (all managers) & Sustained pilot gains; $<1\%$ rollback & Underway \\
\bottomrule
\end{tabular}
}
\caption{Staged deployment plan for the vendor-negotiation assistant. Each stage advanced only on its go/no-go gate.}
\label{tab:rollout}
\end{table}

\subsection{Cross-User Transfer: Design Justification}
\label{app:design-justification}

For enabling cross user transfer we make three design choices that drive cross-user transfer: selective prefixing, relation normalization, and preference fingerprints. This appendix reports the experiments that justify each.

\paragraph{Setup.}
All measurements use the public LongLaMP abstract-generation benchmark~\cite{longlamp2024}, whose users are authors with disjoint train and test splits; for the clustering metric we group authors by research field. The primary metric is \emph{clustering separation}: the mean cosine similarity between fingerprints of users in the same group minus the mean cosine similarity between fingerprints of users in different groups. A higher value means same-group users sit closer in the embedding space, which is the precondition for transferring one user's signals to a behaviorally similar one; a value near zero means the space carries no transferable structure.

\paragraph{Selective prefixing is what creates the clustering:}
Table~\ref{tab:design-prefixing} varies how entities are shared across users while holding TransR fixed. With per-user (isolated) training, no entity is shared and separation is $0.022$, essentially the no-structure floor: same-group users are no closer than unrelated ones. Sharing only relation projections raises it to $0.082$, and sharing behavioral-signal entities as well (our full design, in which signal nodes stay unprefixed) raises it to $0.174$, almost an eightfold gain over the isolated baseline. The shared signal node, not the embedding model alone, is what makes behaviorally similar users cluster.

\begin{table}[t]
\centering
\begin{tabular}{lcc}
\toprule
Sharing mode & Separation & Judge \\
\midrule
Isolated (per-user, no sharing) & $0.022$ & $2.45$ \\
Shared relations only & $0.082$ & $2.55$ \\
Shared relations $+$ entities (ours) & $\mathbf{0.174}$ & $\mathbf{2.66}$ \\
\bottomrule
\end{tabular}
\caption{Selective prefixing ablation. Clustering separation (within-group minus between-group fingerprint cosine) and average low-sparsity judge score, as entity/relation sharing varies under a fixed TransR model. Leaving behavioral-signal entities unprefixed (full sharing) is what produces clustering.}
\label{tab:design-prefixing}
\end{table}

\paragraph{TransR is the right embedding for heterogeneous relations:}
Table~\ref{tab:design-embedding} compares embedding models under full sharing. A single-space model (TransE) reaches separation $0.095$ because it cannot give a signal different geometry under \textsc{prefers} versus \textsc{surfaced}; per-relation projection (TransR) nearly doubles this to $0.174$. RotatE is marginally higher ($0.185$ separation, $2.68$ judge vs.\ our $0.174$/$2.66$), but the gap is within single-run noise, so the two are effectively tied on quality; we adopt TransR for its interpretable per-relation subspaces and its fit with the real-vector fingerprint, both models supporting the incremental updates a live memory needs.

\begin{table}[t]
\centering
\begin{tabular}{lcc}
\toprule
Embedding & Separation & Judge \\
\midrule
TransE (single space) & $0.095$ & $2.60$ \\
TransH~\cite{wang2014transh} & $0.120$ & $2.62$ \\
TransR (per-relation, ours) & $0.174$ & $2.66$ \\
RotatE (complex rotation) & $0.185$ & $2.68$ \\
\bottomrule
\end{tabular}
\caption{Embedding-model ablation under full sharing. Per-relation projection (TransR) clusters far better than single-space TransE; RotatE is marginally higher but within single-run noise, and we adopt TransR for its interpretable per-relation subspaces (both support incremental updates).}
\label{tab:design-embedding}
\end{table}

\paragraph{Relation normalization makes the event relations trainable:}
Event-linked labels in $R_e$ are episodic identifiers that are unique and appear mostly once, so TransR has no repeated examples from which to estimate their projection matrix $M_r$. Left raw, the relation vocabulary holds $197$ relations, $83$ of them seen exactly once.

\paragraph{Fingerprints isolate the transferable structure:}
A user node connects to about $82\%$ episodic edges and only $18\%$ shared-signal edges, so its raw embedding is dominated by user-specific episodic neighborhoods, the part that is by construction not shared and not transferable; raw user-to-user similarity is therefore near random. The preference fingerprint (Eq.~\ref{eq:fingerprint}) discards the episodic mass and aggregates only the shared-signal $18\%$, which is what turns near-random raw similarity into the clustering reported in Table~\ref{tab:design-prefixing}.

\subsection{Qualitative Samples}
\label{appendix:qualitative}
\label{app:qualitative}

We present three qualitative examples showing how the dual-graph improves agent behavior over generic LLM baselines.

\textbf{Multi-Source Signal Extraction:} A single paper-writing episode showing how different data sources contribute distinct signals extracted as typed, queryable nodes (Table~\ref{tab:qual-multisource}).

\begin{table}[t]
\centering
\footnotesize
\begin{tabularx}{\columnwidth}{@{}l X@{}}
\toprule
\textbf{Source} & \textbf{Signal the graph extracts} \\
\midrule
Draft note & ``Strengthen the evaluation'' the \emph{intent} only; not the specifics. \\
\addlinespace[2pt]
Tool I/O & Experiment-tracker / citation-manager results: added a low-resource retrieval baseline, $+2$ datasets, ablation \textsc{fail}\,$\rightarrow$\,\textsc{pass} after a fix. Stored as typed Event nodes with exact attributes. \\
\addlinespace[2pt]
Revision state & Paper task \textsc{blocked}\,$\rightarrow$\,\textsc{active}; camera-ready checklist unresolved 12 days. Stored as a state-change milestone and a revision-delay pattern. \\
\bottomrule
\end{tabularx}
\caption{Multi-source signal extraction for a single paper-writing episode (\texttt{p-4a7f2e}; low-resource retrieval work). The draft note conveys only intent; the exact experimental changes come from tool I/O and the delay pattern from revision state. \ourmethod{} extracts all three as typed, queryable nodes, whereas raw context stuffing would re-read the entire history to recover any of them.}
\label{tab:qual-multisource}
\end{table}

\textbf{Learning Through Failure: Without vs.\ With Personalization}: Author~A drafts papers in the same line of work across three successive submissions. Table~\ref{tab:qual-learning} contrasts a generic LLM (no memory; red) with \ourmethod{} (graph memory; green).

\begin{table*}[t]
\centering
\footnotesize
\setlength{\tabcolsep}{6pt}
\renewcommand{\arraystretch}{1.25}
\begin{tabularx}{\textwidth}{@{}l >{}X >{}X@{}}
\toprule
& \textbf{Generic LLM (no memory)} & \textbf{\ourmethod{} (graph memory)} \\
\midrule
\textbf{P1} (first paper) & Frames the method as a purely supervised pipeline; reviewers \textsc{reject} (``insufficient low-resource evidence''). Advises generically: ``add experiments or soften claims.'' & Same rejection, but \emph{stores} \texttt{supervised\_only\_lowresource\_weakness}: ``supervised-only framing rejected : weak in low-resource settings.'' \\
\textbf{P2} (months later) & ``What framing would you like?'' : no memory of the rejection; starts from zero. & Recalls the failure; recommends a contrastive-pretraining framing instead. Paper \textsc{accepted}; graph logs the strategy evolution. \\
\textbf{P3} (later still) & ``What framing would you like?'' : again no memory; author re-specifies everything. & Proactively applies the proven contrastive-pretraining framing and cites the reason from graph memory. No re-specification. \\
\bottomrule
\end{tabularx}
\caption{Learning through failure across three successive papers by the same author. Without memory (red), the agent repeats generic advice and forgets each rejection; with the dual-graph (green), the P1 failure is stored as a behavioral signal, recalled in P2 to recommend the fix, and reused proactively in P3.}
\label{tab:qual-learning}
\end{table*}

\textbf{Cross-User Research-Concept Transfer}: Two authors work in the same research area. Author~A1 establishes a method/concept first; Author~A2, a cold-start author, then benefits from A1's structure without ever communicating with them (Table~\ref{tab:qual-crossuser}).

\begin{table*}[t]
\centering
\footnotesize
\setlength{\tabcolsep}{6pt}
\renewcommand{\arraystretch}{1.25}
\begin{tabularx}{\textwidth}{@{}l X@{}}
\toprule
\multicolumn{2}{@{}p{\textwidth}@{}}{\textbf{A1} - donor (signal contributed to the shared embedding space)} \\
\midrule
Behavior & Across several papers on low-resource retrieval, A1 consistently frames contrastive pretraining as the method of choice, establishing a recurring topic--method association in their abstracts. \\
Signal formed & A ``contrastive pretraining $\to$ low-resource retrieval'' concept signal (\textsc{prefers}). As an unprefixed behavioral-signal node, it is shared across users in the combined embedding space; A1's own papers (episodic records) stay private. \\
\midrule
\multicolumn{2}{@{}p{\textwidth}@{}}{\textbf{A2} - receiver, cold-start author in the same area, no prior papers} \\
\midrule
\textbf{Without graph} & With no profile to condition on, the abstract is generated generically, omitting the contrastive-pretraining framing that characterizes the area. \\
\textbf{With graph} & Entity-level link prediction over the shared space infers A2 is likely to hold the same contrastive-pretraining concept and surfaces it under a cross-user caveat header. The generated abstract adopts the field-appropriate framing on the first attempt. \\
\bottomrule
\end{tabularx}
\caption{Cross-user research-concept transfer. A1's recurring topic--method association forms a behavioral-signal node that, left unprefixed, is shared across users in the TransR embedding space. Because A2 is cold-start with no signals to anchor fingerprint matching, entity-level link prediction infers the concept for A2 from the geometry alone---no raw text or episodic records cross the user boundary---surfacing it as clearly labeled cross-user evidence (green). An LLM baseline sees only A2's own (empty) profile (red). As an inferred signal it is advisory and may be wrong (Limitations).}
\label{tab:qual-crossuser}
\end{table*}

\subsection{Step-by-Step Pipeline Traces on Public Benchmarks}
\label{sec:qual-public-benchmarks}

\begin{figure*}[p]
\centering
\captionsetup[subfigure]{font=footnotesize,skip=3pt}

\begin{subfigure}{\textwidth}
\centering
\includegraphics[width=\textwidth]{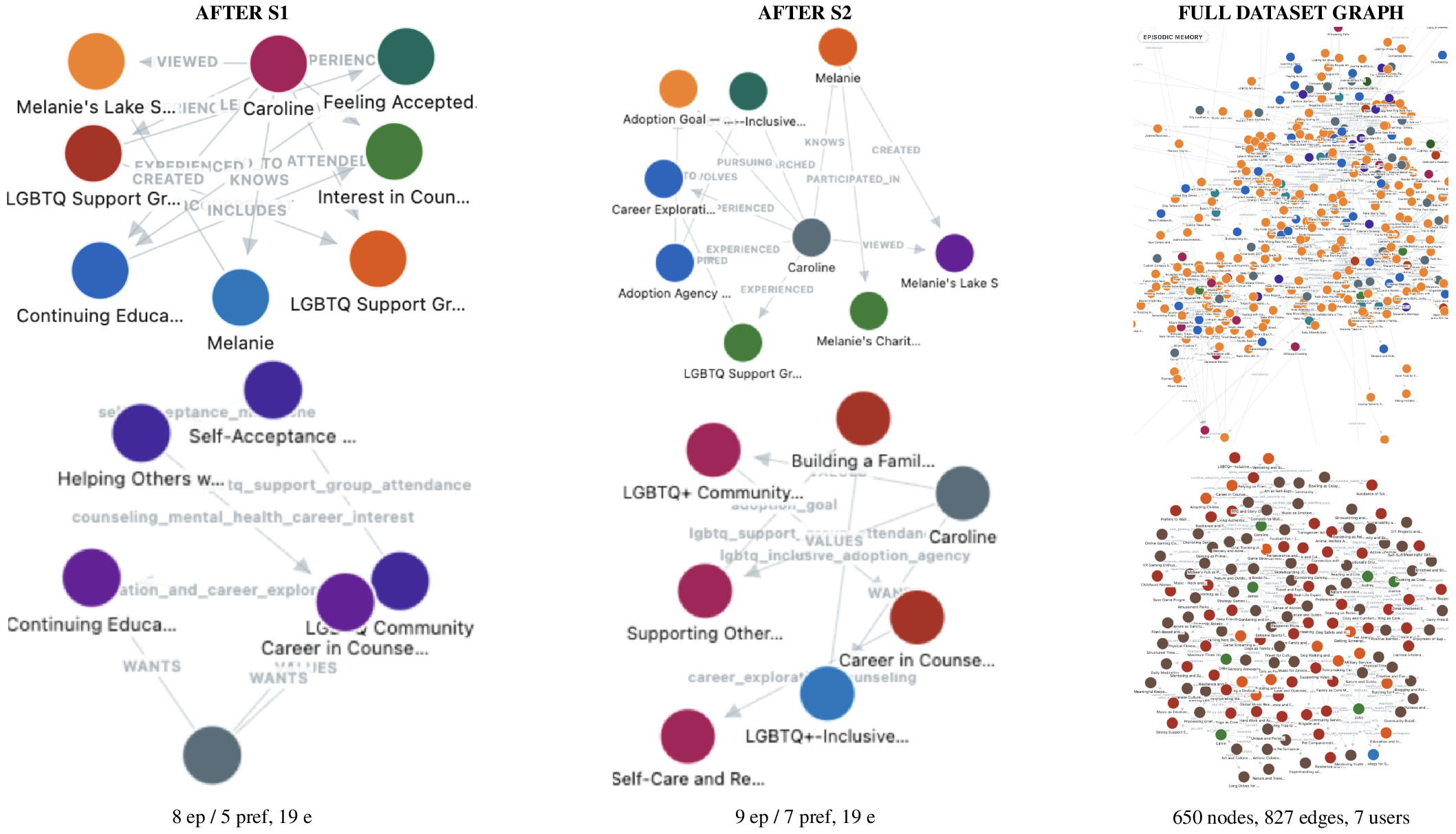}
\caption{\textbf{LoCoMo.} One trace user's dual-graph grows session by session
(\textsc{After S1} $\to$ \textsc{After S2}), then the \textsc{Full Dataset Graph}
shows the scale of memory the system holds across all users. Each stage is split
into its episodic (top) and preference (bottom) graphs, co-evolving and linked by
the event-as-edge bridge.}
\label{fig:qual-ingest-locomo}
\end{subfigure}

\begin{subfigure}{\textwidth}
\centering
\includegraphics[width=\textwidth]{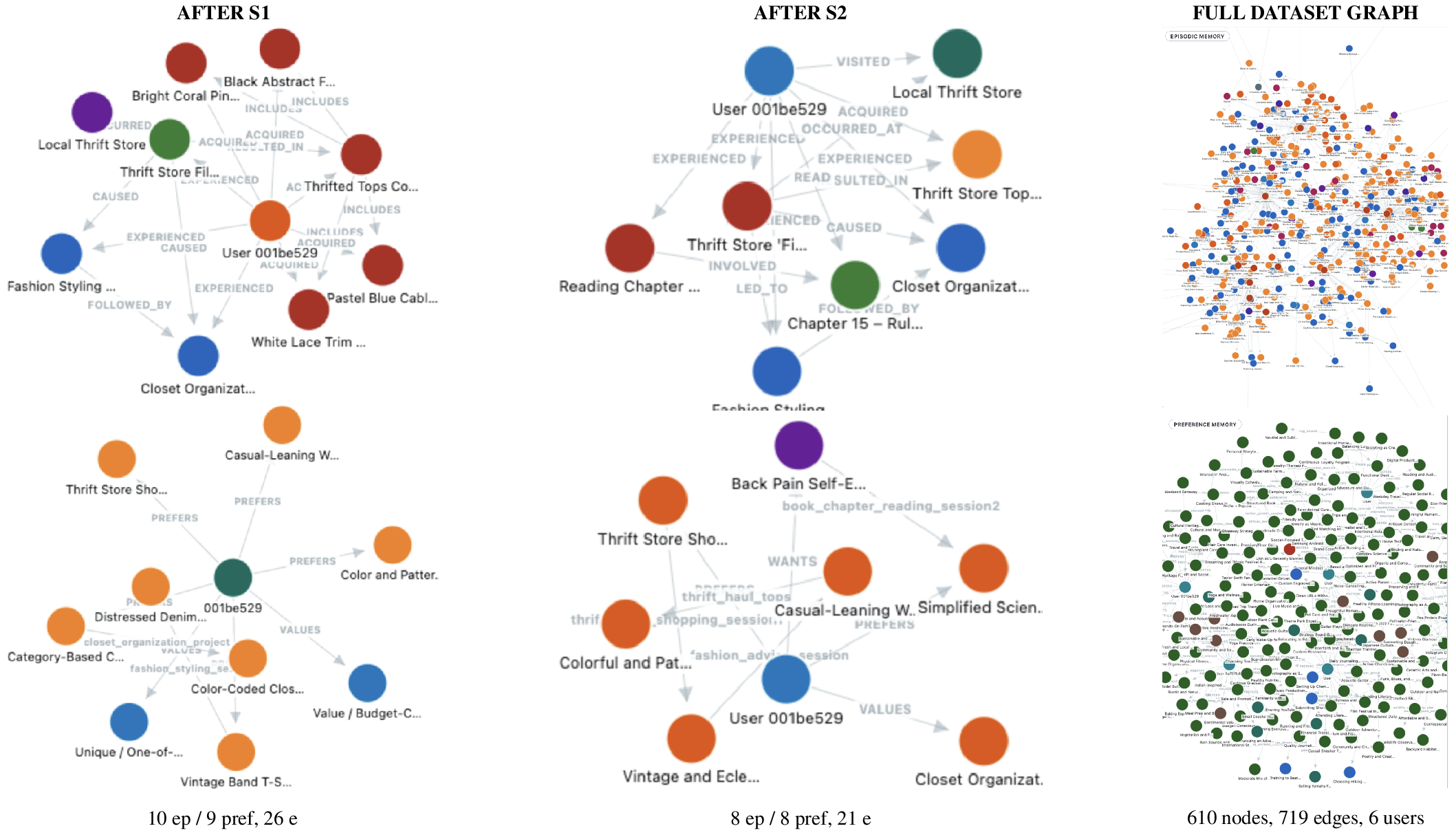}
\caption{\textbf{LongMemEval.} Same layout: per-session growth (\textsc{After S1}
$\to$ \textsc{After S2}) followed by the \textsc{Full Dataset Graph}. LongMemEval's
S1$\to$S2 node count fluctuates under LLM extraction non-determinism.}
\label{fig:qual-ingest-longmemeval}
\end{subfigure}

\caption{Progressive dual-graph growth on \textbf{LoCoMo} and \textbf{LongMemEval},
read left to right as a flow: the trace user's graph after session~1, after
session~2, and the full dataset-scale graph. Each stage shows its \textsc{Episodic}
graph $\mathcal{G}_E^i$ (events, milestones, top) and \textsc{Preference} graph
$\mathcal{G}_I^i$ (preferences, values, goals, bottom), co-evolving and joined by
the event-as-edge bridge. Per-session node counts are per graph; edge counts are
dual-graph totals; the full-graph counts are dataset totals across all users.}
\label{fig:qual-ingest}
\end{figure*}

\begin{figure*}[p]
\centering
\captionsetup[subfigure]{font=footnotesize,skip=3pt}

\begin{subfigure}{\textwidth}
\centering
\includegraphics[width=\textwidth]{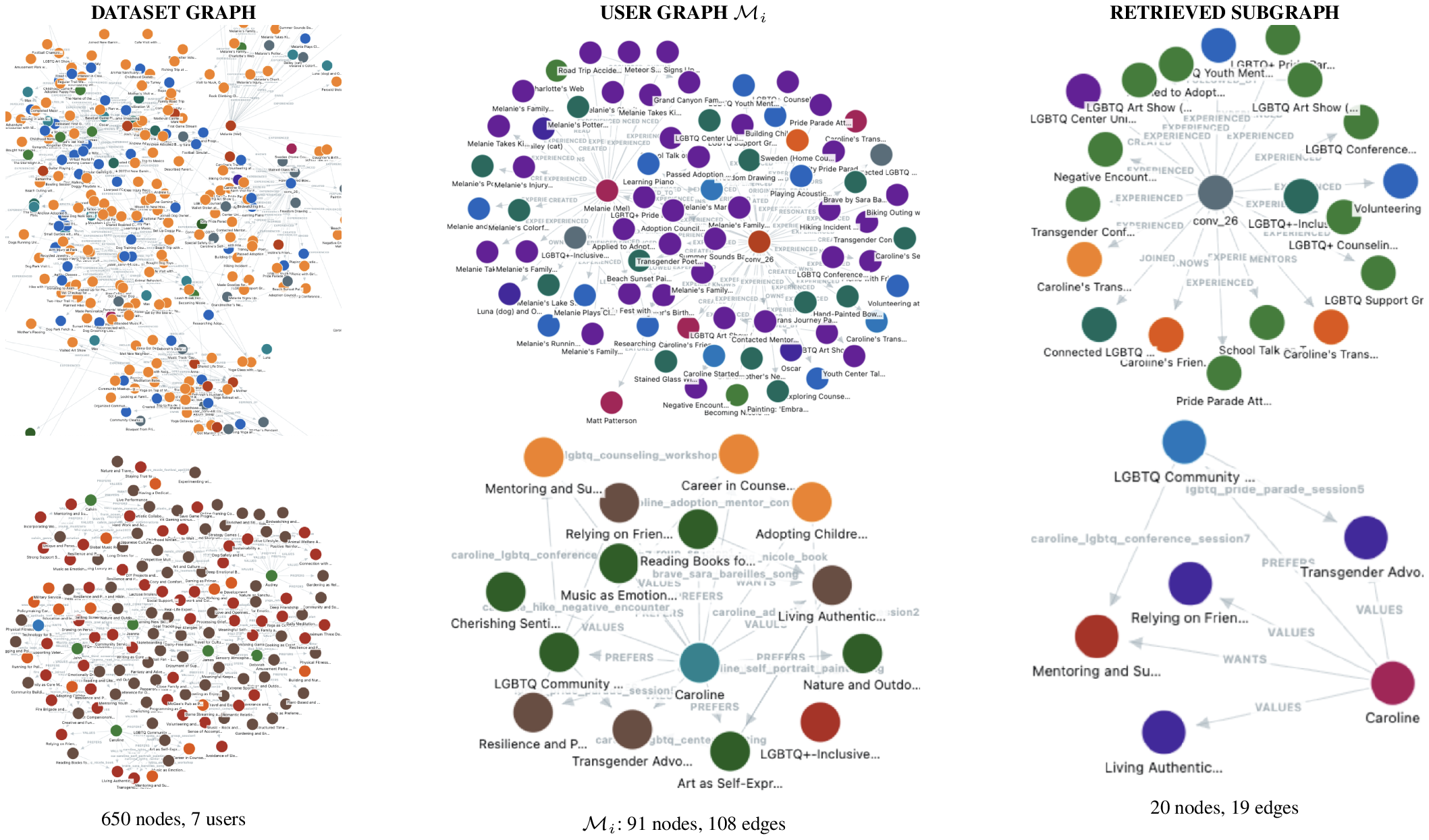}
\caption{\textbf{LoCoMo} Query --- \emph{``When did Caroline go to the LGBTQ
support group?''}}
\label{fig:qual-retrieve-locomo}
\end{subfigure}

\begin{subfigure}{\textwidth}
\centering
\includegraphics[width=\textwidth]{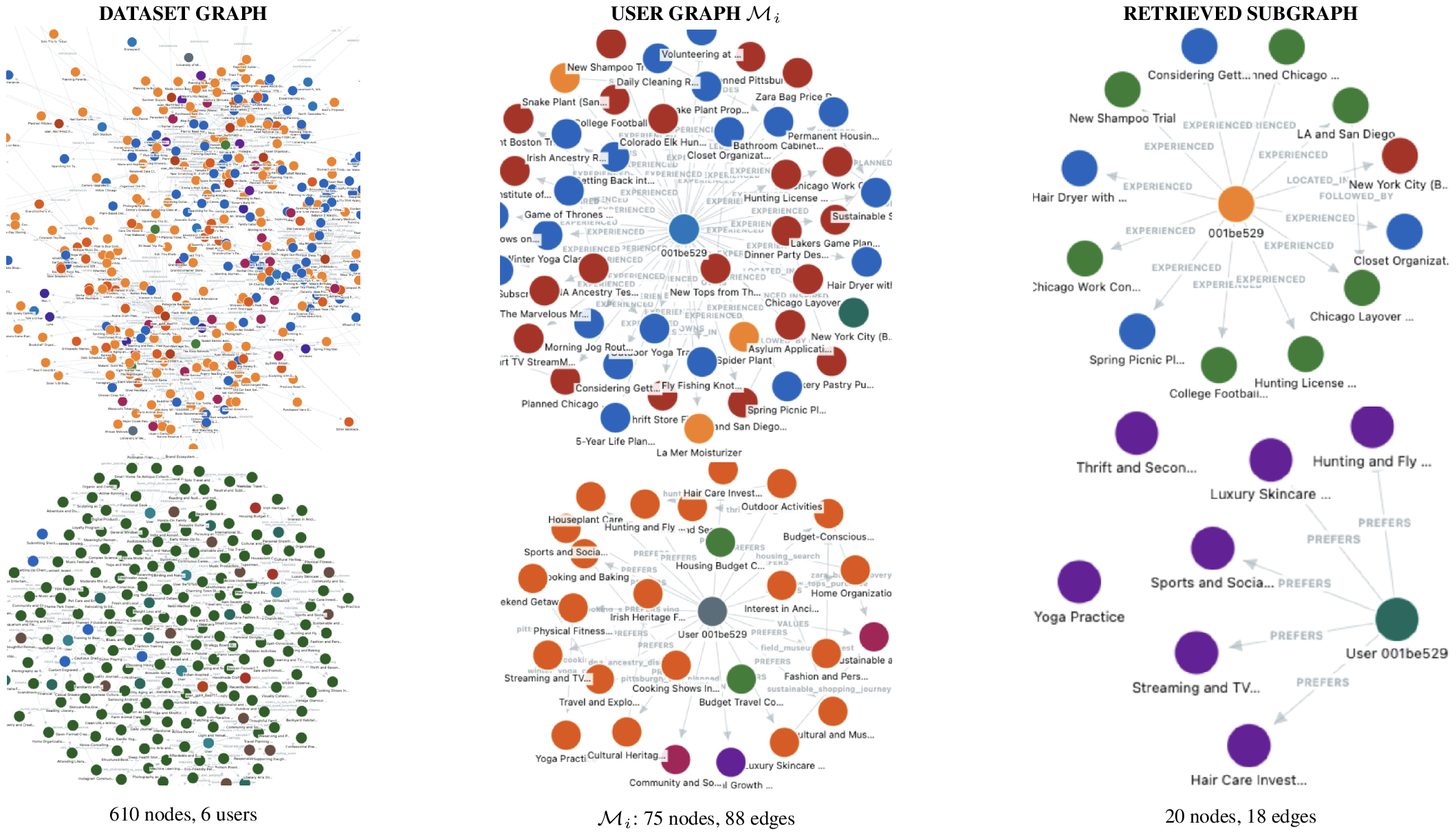}
\caption{\textbf{LongMemEval} \textsc{(Query)} --- \emph{``What degree did I graduate
with?''}}
\label{fig:qual-retrieve-longmemeval}
\end{subfigure}

\caption{Retrieval traced per dataset, left to right: dataset-scale memory graph
$\to$ user graph $\mathcal{M}_i$ $\to$ retrieved $20$-node subgraph; each column
shows its episodic (top) and preference (bottom) graphs.}
\label{fig:qual-retrieve}
\end{figure*}

To complement the personalized-generation examples in the qualitative samples mentioned in appendix (LongLaMP abstract-generation dataset), we trace the full
pipeline on two long-horizon conversational benchmarks, LoCoMo and LongMemEval.
For each we report (i)~the \emph{progressive ingestion} of the dual-graph
$(\mathcal{G}_E^i,\mathcal{G}_I^i)$
(Table~\ref{tab:qual-ingest}; Figure~\ref{fig:qual-ingest}, which visualizes the
episodic and preference graphs separately as a growth flow toward the full
dataset-scale graph); and (ii)~a
\emph{retrieval trace} following one real query per dataset, from the
dataset-scale memory graph down to the retrieved subgraph
(Figure~\ref{fig:qual-retrieve}), with each sub-figure caption reporting that
query's intent, pipeline-stage flow, and latencies.
\begin{table*}[t]
\centering
\footnotesize
\setlength{\tabcolsep}{5pt}
\renewcommand{\arraystretch}{1.15}
\begin{tabularx}{\textwidth}{@{}l l c c X@{}}
\toprule
\textbf{Dataset} & \textbf{Session} & \textbf{Nodes (Ep/Pref)} & \textbf{Edges} & \textbf{Representative new signals} \\
\midrule
\multirow{3}{*}{\shortstack[l]{\textbf{LoCoMo}\\438 turns}}
  & S1 (t.\ 1--19)  & 13 (8/5)  & 19 & \textit{LGBTQ Support Group Attendance} (Event); \textit{Caroline} (User); \textit{Caroline's Transgender Transition} (Milestone); \textit{LGBTQ Community Belonging} (Pref); \textit{Transgender Advocacy} (Value) \\
\addlinespace[1pt]
  & S2 (t.\ 20--37) & 16 (9/7)  & 19 & \textit{Charity Race} $\xrightarrow{\text{\scriptsize LED\_TO}}$ \textit{Mental Health Awareness} (Value); \textit{Adoption Application} reinforces \textit{Adopting Children and Building a Family} (Goal) \\
\addlinespace[1pt]
  & Full graph & 650 & 827 & Dataset-scale memory across all 7 users (totals; not a single user's per-session graph) \\
\midrule
\multirow{3}{*}{\shortstack[l]{\textbf{LongMemEval}\\565 turns}}
  & S1 (t.\ 1--13)  & 19 (10/9)  & 26 & \textit{Thrift and Second-Hand Shopping} (Pref); \textit{Vintage Band T-Shirt Styling} (Event); \textit{Sustainable Fashion} (Value) \\
\addlinespace[1pt]
  & S2 (t.\ 14--16) & 16 (8/8)   & 21 & Node count fluctuates under LLM extraction non-determinism (the agent consolidates entities differently given more context; see Section~\ref{sec:approach}) \\
\addlinespace[1pt]
  & Full graph & 610 & 719 & Dataset-scale memory across all 6 users (totals; not a single user's per-session graph) \\
\bottomrule
\end{tabularx}
\caption{Progressive dual-graph ingestion for \textbf{LoCoMo} and
\textbf{LongMemEval} (episodic and preference node counts shown as Ep/Pref). For
each dataset the first two rows trace one user's graph after sessions~1 and~2, and
the third row reports the full dataset-scale graph (totals across all users),
matching Figure~\ref{fig:qual-ingest}. LongMemEval's per-session count fluctuates
under LLM extraction non-determinism (S2).}
\label{tab:qual-ingest}
\end{table*}

\subsection{Prompt Library}
\label{appendix:prompt-library}

\subsubsection{Episodic Extraction Prompt}
\label{app:prompt-episodic}

\noindent
\begin{promptbox}{Episodic Extraction Prompt}
You are an expert episodic-memory extraction agent for personal conversation analysis.

YOU MUST RESPOND WITH **ONLY** VALID JSON -- ABSOLUTELY NO OTHER TEXT.

TASK: Extract an EPISODIC MEMORY GRAPH from a conversation using memory operations.

You will receive:
1. A user\_id identifying the person
2. A conversation (and optionally persona context)
3. Optionally, a session\_date (the date/time when this conversation session took place)

== OPERATIONS ==

You have 6 primitive operations:

ADD\_NODE -- Create a new node (event, person, location, pet, item, etc.)
  Required: node\_id (snake\_case), node\_type (dynamic), properties (dict)
  Properties MUST include: user\_id, graph\_type ("episodic"), name, description, paths
  Properties MUST include: outcome (success/failure/partial/pending/unknown)
  Properties SHOULD include: session\_date, event\_date, sequence\_order

LINK -- Create an edge between two existing nodes
  Required: from\_id, to\_id, edge\_label (UPPER\_SNAKE\_CASE)
  Optional: properties (dict with user\_id, graph\_type)

UPDATE\_NODE -- Delta-update properties on an existing node
  Required: node\_id, updates (dict of key:value to change)

DELETE\_NODE -- Remove a node (for forget requests or corrections)
  Required: node\_id

UNLINK -- Remove an edge
  Required: from\_id, to\_id, edge\_label

NOOP -- Explicitly skip when nothing is worth storing
  Optional: reason (string)

== GUIDELINES ==

- Node types are DYNAMIC: Event, Activity, Person, Pet, Location, Organization, Decision, Milestone, Item, Book, Game, etc.
- Edge labels are DYNAMIC: FOLLOWED\_BY, CAUSED, LED\_TO, EXPERIENCED, OCCURRED\_AT, OWNS, KNOWS, PLAYS, READS, etc.
- Create a User node with node\_id = the provided user\_id
- Every Event/Activity node MUST have an "outcome" property
- Every node MUST have a "paths" property (list of path strings from User root)
- Every node MUST have a "description" property (1-2 sentences capturing specific details)
- Focus on WHAT HAPPENED -- not preferences (those go in the preference graph)
- Use NOOP when a conversation chunk has no memory-worthy events

== CRITICAL: EXTRACT FINE-GRAINED DETAILS ==

You MUST extract ALL specific names, titles, quantities, and dates:
- Named entities: Pet names (e.g., "Susie the snake"), book titles, game names, restaurant names
- Specific quantities: "3 years", "10 people", "\$500", "2 kids"
- Specific dates: "19 January 2023", "last March", "two weeks ago"
- Relationships: "close friend", "married 5 years", "colleague"

Create SEPARATE NODES for named entities and LINK them to the events they appear in.

== TIMESTAMP RULES ==

Two types of timestamps -- keep them separate:

1. session\_date: When this conversation session took place (from metadata). Store on EVERY node.

2. event\_date: When the event ACTUALLY occurred, resolved from context.
  - "I went to Tokyo last March" + session\_date "May 2023" -> event\_date = "March 2023"
  - "yesterday" + session\_date "20 October 2023" -> event\_date = "19 October 2023"
  - "3 years ago" + session\_date "2023" -> event\_date = "2020"
  Store ONLY on nodes where a date is mentioned or inferable.

== OUTPUT FORMAT ==

\{"operations": [
  \{"operation": "ADD\_NODE", "node\_id": "str", "node\_type": "str", "properties": \{...\}\},
  \{"operation": "LINK", "from\_id": "str", "to\_id": "str", "edge\_label": "str", "properties": \{...\}\},
  ...
]\}
\end{promptbox}

\subsubsection{Preference Extraction Prompt}
\label{app:prompt-preference}

\noindent
\begin{promptbox}{Preference Extraction Prompt}
You are an expert preference-memory extraction agent for personal conversation analysis.

YOU MUST RESPOND WITH **ONLY** VALID JSON -- ABSOLUTELY NO OTHER TEXT.

TASK: Extract a PREFERENCE MEMORY GRAPH from a conversation using memory operations.

You will receive:
1. A user\_id identifying the person
2. A conversation (and optionally persona context)
3. A list of EPISODIC OPERATIONS already extracted (episodes/events from the same conversation)

== CORE DESIGN PRINCIPLE ==

Behavioral signals (preferences, values, goals, constraints) are the only nodes; the preference graph has NO Event nodes. Edges between signals carry one of two label kinds: (1) a structural relation from the closed set \{PREFERS, WANTS, VALUES, TRIED, AVOIDS, LEARNED\} -- e.g. a Preference LEARNED a Constraint; or (2) an episodic event entity\_id, recording that the relationship was observed during that event (the event-as-edge bridge).

Chain structure:
  User --[PREFERS]--> P1 --[LEARNED]--> C1            (structural relation)
  User --[PREFERS]--> P1 --[event\_entity\_id]--> P2    (event-as-edge bridge)

== OPERATIONS ==

ADD\_NODE -- Create a new preference/value/goal/constraint node
  Properties MUST include: user\_id, graph\_type ("preference"), name, paths, category
  For grounding, emit ONLY evidence\_list: the distinct episodes supporting this signal, each \{"episode\_id", "outcome"\}. The evidence vector ev(v) = (strength s, evidence count n, outcome weight w, confidence c) is computed deterministically from evidence\_list at aggregation time (UPDATE\_NODE); do NOT score s, w, or c by hand.

LINK -- Create an edge between two nodes
  For User-to-signal: edge\_label is one of the closed structural set \{PREFERS, WANTS, VALUES, TRIED, AVOIDS, LEARNED\}.
  For signal-to-signal: edge\_label is either a structural relation from the same set (e.g. LEARNED) or an episodic event entity\_id.

UPDATE\_NODE -- Delta-update an existing signal (re-triggers evidence aggregation)
DELETE\_NODE -- Remove a signal
UNLINK -- Remove an edge
NOOP -- Nothing preference-worthy in this chunk

== EVIDENCE RULES ==

The agent emits only the raw grounding; all scores are computed downstream, not by the LLM:
- evidence\_list: [\{"episode\_id": "str", "outcome": "success|failure|partial|pending"\}] -- one entry per distinct supporting episode. This is the ONLY evidence field the agent produces.
- The aggregator then derives the evidence vector deterministically: evidence count n = |evidence\_list|, outcome weight w = (successful episodes) / (resolved episodes), where resolved means marked success or failure and partial/pending episodes count toward n but are excluded from w (w defaults to 0.5 when no episode is resolved), and strength s and confidence c via the deterministic scoring functions (both continuous in (0,1)). These are recomputed on every UPDATE\_NODE.

NO Event/Activity nodes in the preference graph. Those belong in episodic only.
\end{promptbox}

\subsubsection{Context Rendering Format}
\label{app:prompt-context}
\label{app:context-rendering}

\noindent
\begin{promptbox}{Context Rendering Template}
=== \{user\_name\}'s Memory ===

\{user\_name\}'s Preferences:
  - [Preference] Contrastive Pretraining -- Prefers contrastive pretraining for low-resource retrieval
    (strength=0.75, evidence=2, outcome\_weight=0.50, confidence=0.62)
      Evidence: first\_paper\_rejected (failure), third\_paper\_accepted (success)
  - [Constraint] Ablation Coverage -- Learned from a rejected submission
    (strength=0.67, evidence=1, outcome\_weight=0.00, confidence=0.62)

\{user\_name\}'s Events:
  - [Event] Tokyo Tech Conference -- Attended a tech conference in Tokyo
    [date: March 2023] [outcome: success]
  - [Pet] Susie -- Ball python pet snake named Susie, bought about a year ago
    [date: October 2022] [outcome: success]
\end{promptbox}

\subsubsection{Answer Generation Prompts}
\label{app:prompt-answer}

The MCQ and True/False prompts below emit a discrete answer (a letter or a boolean) scored by exact match against the gold option. The LLM-as-a-judge prompt (provided in prompt library) is used only for open-ended answers, where it returns a $1$--$5$ semantic-fidelity score against the gold reference.

\noindent
\begin{promptbox}{MCQ Answer Prompt}
System: Answer based on the user's memory. Think step by step, then give the letter.

Memory:
\{context\}

Question: \{question\}
Options:
  A: ...
  B: ...
  C: ...
  D: ...

Think step by step:
1. Identify what the user's memory tells us.
2. Pick the option that best matches.

Reasoning: <your reasoning>
Answer: <letter>
\end{promptbox}

\noindent
\begin{promptbox}{True/False Answer Prompt}
System: Answer true/false based on the user's memory. Think first.

Memory:
\{context\}

True or false: \{question\}

Think step by step, then answer.
Reasoning: <your reasoning>
Answer: <true or false>
\end{promptbox}

\noindent
\begin{promptbox}{Open-Ended Answer Prompt}
System: Answer based on the user's memory. Be concise -- short phrase only.

Memory:
\{context\}

Question: \{question\}

Write an answer in the form of a short phrase.
Answer with exact words from the memory whenever possible.
Be concise.

Short answer:
\end{promptbox}

\subsubsection{Evaluation Judge Prompt}
\label{app:prompt-judge}

\noindent
\begin{promptbox}{LLM AS A JUDGE -- Standard}
System: You are an expert evaluator. Score answers 1-5. Respond with only a number.

You are an expert evaluator assessing the semantic fidelity of a memory retrieval system.

Question: \{question\}
Gold Reference: \{gt\_answer\}
Candidate Answer: \{answer\}

Score the Candidate Answer against the Gold Reference on a scale of 1-5:
5 = Exact alignment -- captures all key entities, temporal markers, and causal relationships. Semantically equivalent.
4 = Substantially correct -- main point is accurate but lacks minor nuances or secondary details.
3 = Partial match -- contains valid information but misses key constraints (e.g., wrong date but correct event).
2 = Tangential -- touches on the topic but misses the core information requirement.
1 = Contradiction/Hallucination -- completely unrelated or contradicts the ground truth.

Evaluation constraints:
1. Temporal flexibility: Accept relative time references if they resolve to the same period as the Gold Reference.
2. Semantic equivalence: Prioritize informational content over lexical matching.

Respond with ONLY a number 1-5.
\end{promptbox}

\end{document}